\documentclass[10pt,twocolumn,letterpaper]{article}

\usepackage{cvpr}              
\definecolor{cvprblue}{rgb}{0.21,0.49,0.74}
\usepackage[pagebackref,breaklinks,colorlinks,allcolors=cvprblue]{hyperref}
\usepackage{algorithm}
\usepackage{algpseudocode}
\usepackage{multirow}
\usepackage{amsmath}
\usepackage{amssymb}
\usepackage{booktabs}
\usepackage{xcolor}
\def\paperID{38450} 
\def\confName{CVPR}
\def\confYear{2026}

\title{GeoFusion-CAD: Structure-Aware Diffusion with Geometric State Space for Parametric 3D Design\thanks{This work was supported in part by the National Natural Science Foundation of China under Grant 62276232 and the Key Program of Natural Science Foundation of Zhejiang Province under Grant LZ24F030012.}}

\author{
    Xiaolei Zhou$^1$ \quad 
    Chuangjie Fang$^1$ \quad 
    Jie Wu$^2$ \quad 
    Jingyi Yang$^1$ \quad 
    Boyi Lin$^1$ \quad 
    Jianwei Zheng$^{1, \ast}$ \\
    $^1$Zhejiang University of Technology \quad $^2$Hangzhou International Innovation Institute, Beihang University \\
    {\tt\small \{jimz, fangcj, yjy, linboe, zjw\}@zjut.edu.cn, jiewu@buaa.edu.cn} 
}

\let\oldtwocolumn\twocolumn
\renewcommand\twocolumn[1][]{%
    \oldtwocolumn[{#1}{
    \begin{center}
           \includegraphics[width=\textwidth]{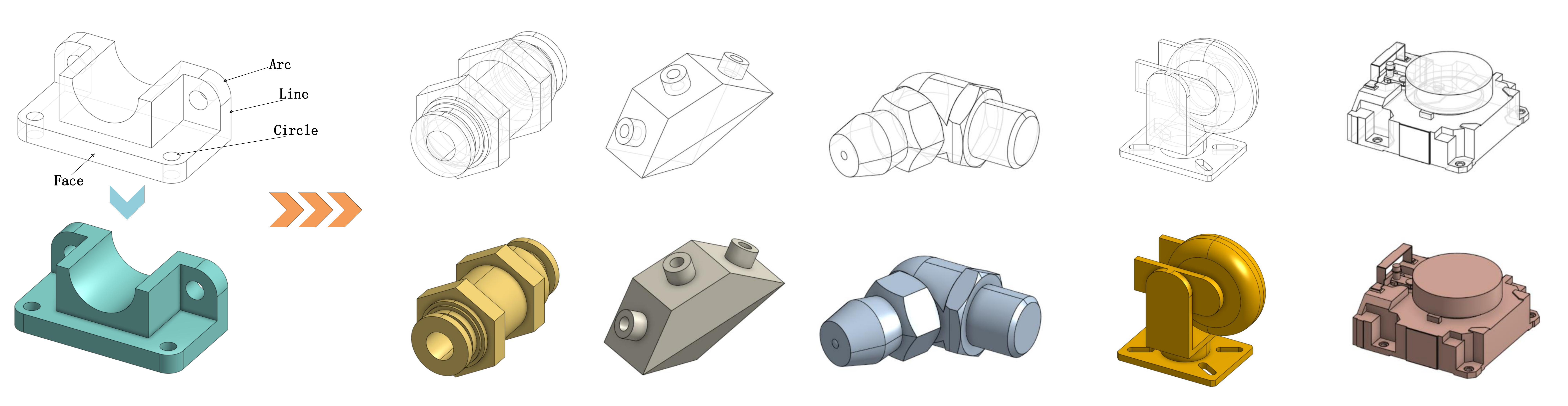}
           \captionof{figure}{{\bf GeoFusion-CAD for 3D modeling}, which learns Sketch-Extrusion (Top) to reconstruct CAD model (Bottom).}
           \label{fig:firstpage}
        \end{center}
    }]
}

\begin{document}
\maketitle
\let\thefootnote\relax\footnotetext{$^\ast$Corresponding author.}

\begin{abstract}
Parametric Computer-Aided Design (CAD) is fundamental to modern 3D modeling, yet existing methods struggle to generate long command sequences, especially under complex geometric and topological dependencies. 
Transformer-based architectures dominate CAD sequence generation due to their strong dependency modeling, but their quadratic attention cost and limited context windowing hinder scalability to long programs. 
We propose GeoFusion-CAD, an end-to-end diffusion framework for scalable and structure-aware generation, as illustrated in Fig. \ref{fig:firstpage}. 
Our proposal encodes CAD programs as hierarchical trees, jointly capturing geometry and topology within a state-space diffusion process. 
Specifically, a lightweight G-Mamba block models long-range structural dependencies through selective state transitions, enabling coherent generation across extended command sequences. 
To support long-sequence evaluation, we introduce DeepCAD-240, an extended benchmark that increases the sequence length ranging from 40 to 240 while preserving sketch–extrusion semantics from the ABC dataset. 
Extensive experiments demonstrate that GeoFusion-CAD achieves superior performance on both short and long command ranges, maintaining high geometric fidelity and topological consistency where Transformer-based models degrade. 
Our approach sets new state-of-the-art scores for long-sequence parametric CAD generation, establishing a scalable foundation for next-generation CAD modeling systems. Codes and datasets are attached and will be released at GitHub.
\end{abstract}

\section{Introduction}

Computer-aided design (CAD) has become a cornerstone of modern mechanical, industrial, and product design, serving as the procedural foundation for creating parameterized 3D geometries and assemblies \cite{robertson1993cad,li2024cad}.  
Among various paradigms, two complementary approaches have dominated recent research: \textit{Sketch–Extrusion (SE)} and \textit{Boundary Representation (B-Rep)}.  
SE models construct solids by sequentially generating 2D sketches—composed of primitives such as lines, arcs, and circles—and applying operations like extrusion, revolution, or sweep to form 3D finals \cite{xu2021inferring,wu2021deepcad,ren2022extrudenet,li2023secad}.  
This procedural formulation preserves parametric constraints and design intent, supporting editable and interpretable modeling workflows.  
In contrast, B-Rep represents solids through topologically continuous boundaries such as NURBS or B-spline surfaces \cite{xu2023hierarchical,xu2024brepgen}, offering complete geometric and topological consistency.  
While the sketch–extrusion branch emphasizes hierarchical parametric reasoning, B-Rep focuses on globally consistent topology and geometric rigor.  
The fundamental divergence between these paradigms, i.e., procedural versus boundary-based representations, makes generative CAD modeling particularly challenging, as it requires unifying design intent with structural continuity.

Currently, transformer-based architectures dominate CAD sequence generation due to the appealing performance gains \cite{wu2021deepcad,xu2022skexgen,xu2023hierarchical,xu2024brepgen}. The core merits arise from the mechanism of self-attention, which enables long-range dependency learning. However, the quadratic time and memory complexity $\mathcal{O}(L^2 d)$ quickly becomes prohibitive as CAD programs scale to hundreds of commands.  
To mitigate this cost, most current solutions naturally divide long sequences into short fragments or train reconstruction and generation stages separately in latent space.  
However, such decomposition breaks end-to-end optimization, leading to feature misalignment and information loss.  
Moreover, global attention treats all tokens equally, ignoring the hierarchical organization of CAD data.  
In practice, CAD entities such as sketches, faces, and extrusions follow strict topological dependencies, where local features must remain consistent within global design contexts.  
Uniform attention thus dilutes these local relations, causing discontinuities in geometric reasoning and reduced coherence in long, structured CAD sequences.  
Consequently, transformer-based methods struggle to balance global context modeling with local structural fidelity, leaving room for further improvements.

\begin{figure}[t]
    \centering
    \includegraphics[width=1\linewidth]{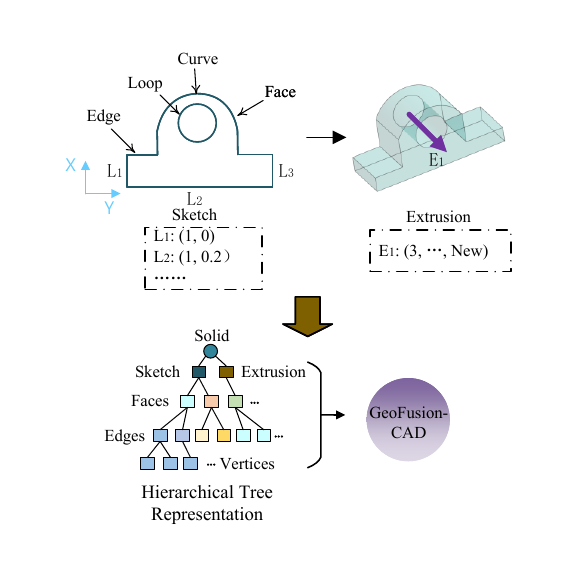}
    \vspace{-0.8cm}
    \caption{
        CAD construction pipeline and hierarchical representation. Sequential Sketch and Extrusion operations form a 3D solid, which is encoded by GeoFusion-CAD into a hierarchical tree capturing geometry and topology.}
    \label{fig:enter-label}
\end{figure}

To overcome the limitations, we propose \textbf{GeoFusion-CAD}, an end-to-end diffusion framework that leverages geometric state-space modeling for efficient and structure-aware CAD generation.  
Instead of relying on dense global attention, we introduce a linear-time G-Mamba diffusion encoder that captures geometric dependencies through selective state transitions with complexity $\mathcal{O}(Ld)$.  
This design combines the scalability of state-space models with the hierarchical awareness needed for procedural reasoning.  
As illustrated in Fig.~\ref{fig:enter-label}, GeoFusion-CAD represents CAD programs as hierarchical trees, where different nodes encode geometric primitives and topological relations individually.  
Information is propagated through G-Mamba blocks that fuse local geometry and global topology in a unified latent space, while a diffusion-based denoising objective ensures stable optimization and robust generalization to long sequences. Together, these innovations enable efficient, interpretable, and topology-consistent CAD generation at scale. To evaluate scalability and generalization, we further extend the DeepCAD dataset to create DeepCAD-240, expanding the command length from 40 to 240 while preserving the original sketch–extrusion semantics.  
This new benchmark provides a challenging testbed for studying long-range dependency modeling, computational efficiency, and geometric consistency in procedural CAD generation.

\noindent{Our main contributions are as follows.}
\begin{itemize}
  \item We propose GeoFusion-CAD, an end-to-end diffusion framework that integrates G-Mamba blocks to achieve efficient and structure-aware CAD sequence modeling.
  \item We design a hierarchical tree representation that explicitly encodes the geometric, parametric, and topological dependencies of Sketch–Extrusion modeling, enabling coherent and interpretable CAD reconstruction.
  \item We construct DeepCAD-240, an extended benchmark dataset with command sequences up to 240 steps, verifying that GeoFusion-CAD achieves scalable and accurate CAD generation.
\end{itemize}

\section{Related Work}
\subsection{Attention-based CAD Modeling}
Attention mechanisms have been widely applied to CAD reconstruction and geometric modeling. 
Recent approaches fit parametric curves \cite{cherenkova2023sepicnet,wang2020pie,mallis2023sharp} or surfaces \cite{le2021cpfn,guo2022complexgen,sharma2020parsenet} from point clouds and B-Rep data \cite{jayaraman2022solidgen,xu2024brepgen}, while Constructive Solid Geometry (CSG) methods \cite{du2018inversecsg,friedrich2019optimizing,kania2020ucsg,geometry2022neural,yu2023d,xu2025laboring,xu2026trt} combine primitive solids via Boolean operations to approximate coarse procedural histories. 
However, attention-based architectures suffer from quadratic complexity, making them inefficient for long and structured CAD programs. 
Moreover, global self-attention treats all tokens uniformly, overlooking the hierarchical dependencies among sketches, features, faces, edges, and vertices. 
This uniformity dilutes local geometric relations and weakens long-range structural coherence, limiting the scalability of systems such as DeepCAD \cite{wu2021deepcad} and MultiCAD \cite{ma2023multicad}. 
Although recent state-space models such as Mamba \cite{gu2023mamba} achieve linear-time computation, their rigid sequential scanning restricts the modeling of hierarchical topological dependencies. 
These limitations motivate the need for a unified framework that efficiently captures geometric hierarchy, long-range dependencies, and procedural consistency.


\subsection{CAD as Language enc-dec Modeling}
Viewing CAD as a structured language \cite{ganin2021computer,seff2022vitruvion}, 
Sketch--Extrusion programs can be tokenized into operation and parameter sequences and decoded autoregressively. 
transformer encoder-decoder architectures \cite{vaswani2017attention} provide powerful sequence modeling capabilities and have been widely adopted in CAD systems such as DeepCAD \cite{wu2021deepcad}, SkexGen \cite{xu2022skexgen}, CAD-SIGNet \cite{khan2024cad}, and MultiCAD \cite{ma2023multicad}. 
SketchGen \cite{para2021sketchgen}, PolyGen \cite{nash2020polygen}, and TurtleGen \cite{willis2021engineering} further extend language modeling to 2D sketches, 3D meshes, and graph-structured geometry using pointer networks \cite{vinyals2015pointer} and hierarchical decoders. 

Despite these advances, transformer-based encoder--decoder models remain limited for long procedural CAD sequences. 
Quadratic self-attention limits scalability, while the lack of hierarchical inductive bias makes maintaining topological consistency across sketches, faces, and solids challenging. 
Diffusion-based extensions such as BrepGen \cite{xu2024brepgen} improve stability but continue to struggle with long-range coherence and multi-stage optimization. 
Collectively, existing approaches demonstrate the promise of CAD as language, yet expose the inefficiency and structural limitations of attention-based architectures in modeling long, hierarchical CAD programs.

\subsection{Diffusion Models for Geometry Generation}
Diffusion models have shown strong potential for generating structured geometry with known or partial topology. 
HouseDiffusion \cite{shabani2023housediffusion} generates architectural layouts under connectivity constraints, while CAGE \cite{liu2024cage} synthesizes articulated 3D structures guided by topology. 
PolyDiff \cite{alliegro2023polydiff} extends diffusion to unstructured polygonal meshes but lacks hierarchical semantics.
Technically, applying diffusion to parametric CAD remains challenging due to long, interdependent command sequences and hierarchical dependencies. 
Existing pipelines often rely on multi-stage or latent training for stability, increasing cost and reducing parametric consistency. 
In contrast, our approach integrates diffusion with structured representation learning, applying it directly on hierarchical CAD sequences through geometric state-space modeling. 
This enables coherent modeling of geometry and topology while improving scalability for long sequences, as demonstrated on extended DeepCAD-240.

\section{CAD Data Geometry}
A CAD solid is typically composed of geometric primitives such as lines, arcs, and circles, combined with parametric extrusion operations as illustrated in Fig.~\ref{fig:enter-label}. The central challenge in CAD data modeling lies in representing both geometric precision and topological consistency in a unified form. To address this challenge, we design a hierarchical tree representation that organizes CAD geometry across multiple semantic levels, as shown in the bottom segment of Fig.~\ref{fig:enter-label}. Each node in the tree encodes both geometric attributes and topological relations. Unlike previous methods~\cite{xu2024brepgen,xu2023hierarchical} that replicate nodes to represent shared edges, our structure preserves connectivity without duplication, maintaining the complete procedural history of the CAD design. The root node corresponds to the overall solid model, while child nodes represent sketches and extrusion operations. From top to bottom, the hierarchy progresses through three levels—operations or faces, edges or extrusion depths, and vertices—capturing both global design intent and fine-grained local geometry. This unified structure reconciles SE representations into a consistent format suitable for diffusion-based generation. Complete implementation details are provided in the supplementary material.

\subsection{Sketch Representation}
A sketch node is encoded as a feature vector that combines geometric and positional parameters. 
As shown in Fig.~\ref{fig:enter-label}, each sketch consists of one or more faces, which are delineated by loops. 
A loop is a closed trajectory that can be either a single primitive (e.g., a circle) or a chain of multiple curves (e.g., lines and arcs). 
Each curve is parameterized by discrete 2D coordinates $(p_x, p_y)$ that define endpoints or control points. 
Termination symbols $e_c$, $e_l$, $e_f$, and $e_s$ mark the end of curves, loops, faces, and sketches, respectively, thereby encoding hierarchical boundaries within the sequence. 
All sketches are represented as token sequences $\mathcal{S}$, capturing both geometric features and structural dependencies.

\begin{figure*}[t]
\centering
\includegraphics[width=1\linewidth]{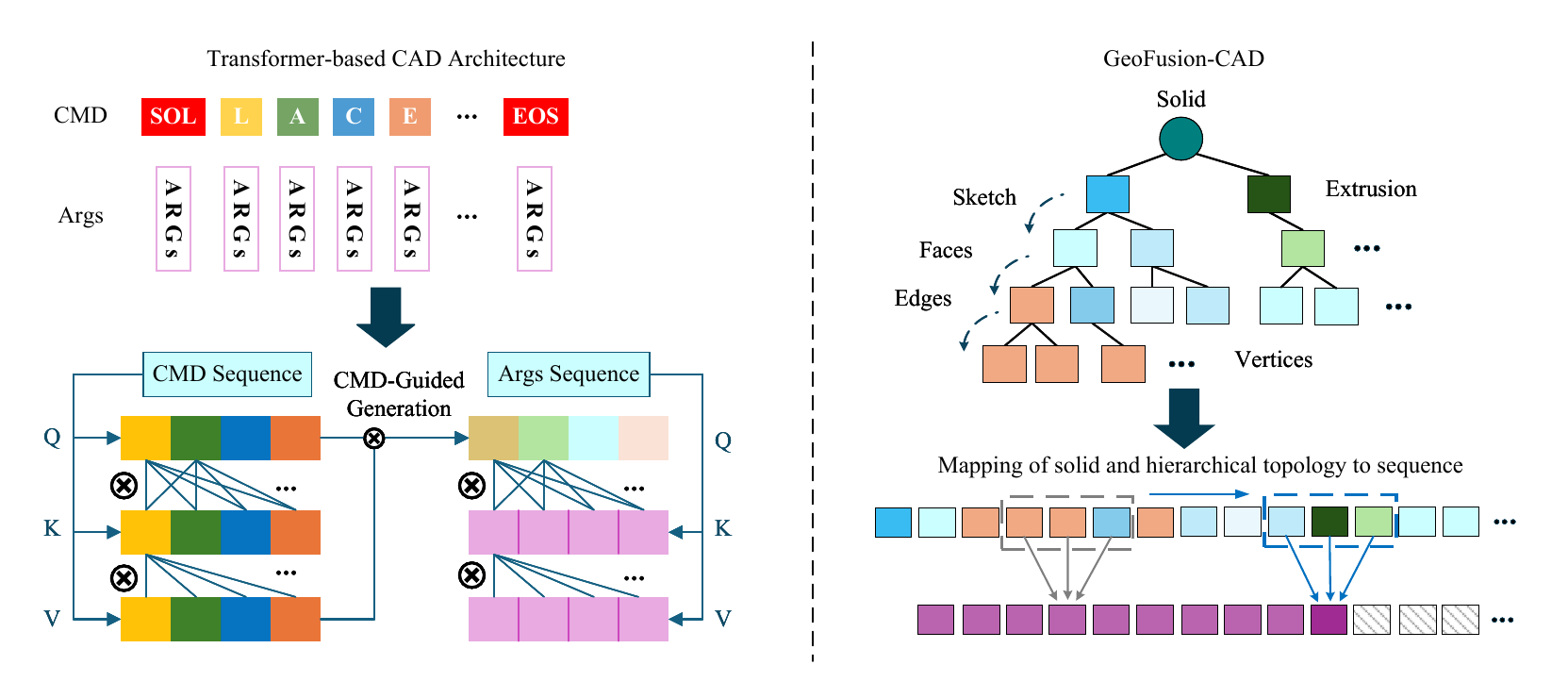}
\vspace{-0.3cm}
\caption{
\textbf{Comparison between transformer-based CAD architecture and ours.}
(Left) The transformer-based method trains command and argument sequences separately. 
(Right) Our GeoFusion-CAD models the hierarchical CAD topology with G-Mamba blocks, mapping solid and hierarchical topology into a unified sequence.
}
\label{fig:vs}
\vspace{-0.2cm}
\end{figure*}

\subsection{Extrusion Representation}
The extrusion operation defines how a 2D sketch is transformed into a 3D solid volume. 
Following~\cite{khan2024cad}, we represent extrusion parameters using orientation angles $(\theta, \phi, \gamma)$ and translation offsets $(\tau_x, \tau_y, \tau_z)$ with respect to a global coordinate system. 
The parameter $\sigma$ specifies the scale of the sketch, while $(d_+, d_-)$ denote the extrusion distances along the normal and opposite directions. 
An additional categorical token $\beta$ indicates the extrusion type (\textit{new, cut, join, intersect}), and $e_e$ marks the end of the extrusion sequence. 
All parameters are converted into discrete tokens $\mathcal{E}$, providing a compact and structured representation of extrusion transformations.

In addition to sketch tokens $\mathcal{S}$ and extrusion tokens $\mathcal{E}$, we append \texttt{pad} and \texttt{cls} tokens to indicate padding and sequence boundaries. 
Together, these elements form the complete sketch–extrusion sequence $\mathcal{C}$, which preserves both geometric design history and topological coherence. 
Following~\cite{wu2021deepcad,khan2024cad}, the design process is represented as a sequence $\mathcal{C} = \{\mathcal{C}_j\}_{j=1}^{n_s}$, 
where each design step $\mathcal{C}_j = \{t_k\}_{k=1}^{n_j}$ contains $n_j$ tokens $t_k \in [0, 1, ..., d_t]$, with $d_t$ denoting the tokenization interval.


\subsection{DeepCAD-240 Dataset}
We construct an extended benchmark, DeepCAD-240, based on the original DeepCAD dataset~\cite{wu2021deepcad}.  
It expands the max command length from 60 up to 240, while retaining the sketch–extrusion semantics and tokenization protocol.  
DeepCAD-240 introduces richer hierarchical dependencies and longer-range geometric contexts, 
providing a more challenging scenario for the evaluation of scalable CAD generation. Due to page limitations, further construction details are provided in the supplementary materials.

\begin{figure*}[t]
	\centering
	\includegraphics[width=1\linewidth]{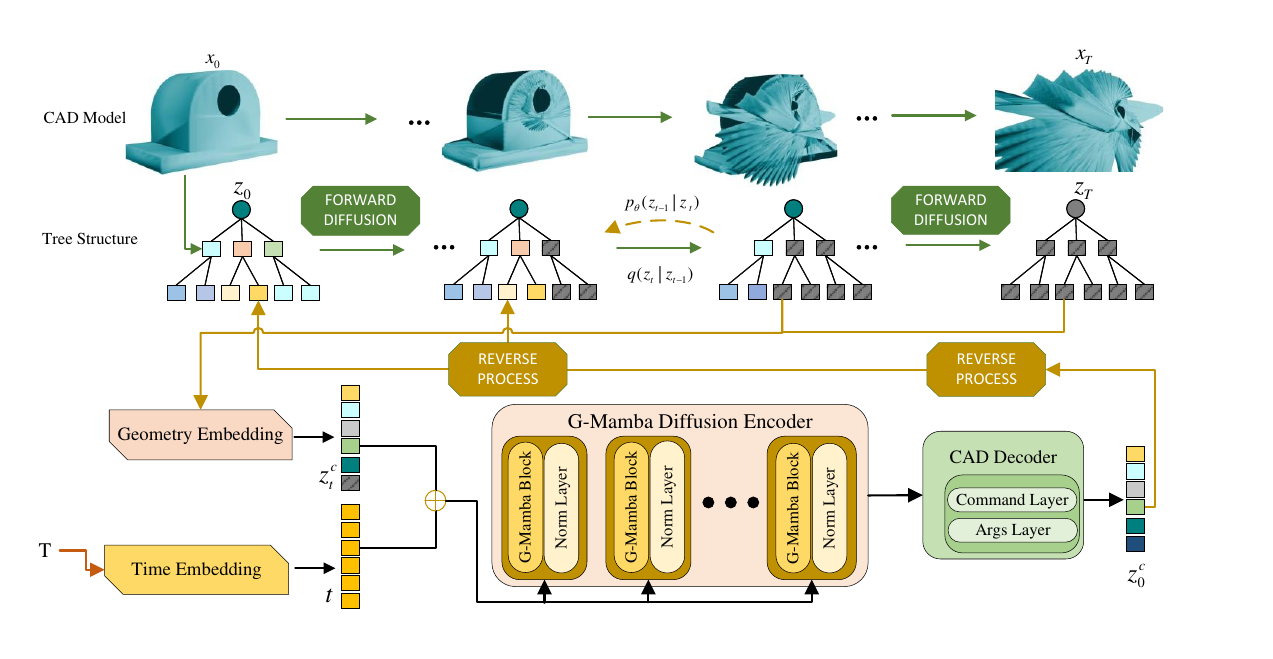}
     \vspace{-0.4cm}
	\caption{\textbf{Overview of GeoFusion-CAD.} 
	The framework consists of two embedding layers, a G-Mamba diffusion encoder, and a CAD decoder. 
	Each parent node conditions its child node during the reverse diffusion process, enabling structured and coherent generation.}
	\label{fig:2}
    \vspace{-0.2cm}
\end{figure*}

\section{Methodology}


\subsection{Problem and GeoFusion-CAD Architecture}

Most existing CAD generation frameworks~\cite{xu2023hierarchical,xu2022skexgen,wu2021deepcad} adopt transformer-based architectures, which face both computational and structural limitations. The quadratic complexity $\mathcal{O}(L^2d)$ of self-attention makes long CAD sequences computationally expensive, while stage-wise pipelines that separately train command and argument tokens result in optimization inconsistency and information fragmentation. As illustrated in Fig.~\ref{fig:vs}, transformer-based methods decompose the procedural CAD sequence into independent command and argument streams. This separation neglects the tight coupling between operation semantics and geometric parameters, breaking the continuity of hierarchical reasoning across solids, sketches and extrusions. Moreover, the global self-attention treats all tokens equally, which dilutes the local geometric dependencies necessary for preserving hierarchical topology. In practice, this means that a local geometric element from sketches (e.g., line, arc, and circle) must attend to all tokens, including extrusion operations, causing context dilution and weakening topological consistency.

To overcome these limitations, we propose GeoFusion-CAD, a single-stage diffusion framework based on geometric state-space modeling for efficient and topology-aware CAD generation. 
By replacing the quadratic self-attention with a linear-time G-Mamba diffusion encoder, our method models geometric dependencies through selective state transitions with complexity $\mathcal{O}(Ld)$, achieving both scalability and topological fidelity. 
As illustrated in Fig.~\ref{fig:vs}, our hierarchical design encodes the CAD tree (solids, sketches, faces, edges, and vertices) into a unified sequence representation, enabling cross-level information fusion and preserving local-to-global geometric consistency. 
The full architecture consists of a Time Embedding for diffusion timesteps, a Geometry Embedding for tokenized CAD sequences, a G-Mamba diffusion encoder for hierarchical feature denoising, and a CAD decoder that reconstructs coherent parametric solids from the refined latent sequence.

\subsection{Geometry Embedding}
 Given an input CAD sequence $\mathcal{C}$ consisting of geometric sequence $\mathcal{S}$ and extrusion operation $\mathcal{E}$, we follow \cite{khan2024cad} that considers the sketch coordinates $p_x$ and $p_y$ as a single 2-dimensional token $(p_x,p_y)$. Note that $\mathcal{C} = \{{t_i}\}_{i=1}^{n_{ts}} \in [0, 1, ..., d_t]^{n_{ts}}$, with $d_t$ defining the tokenization interval and $n_{ts} = \sum_{j=1}^{n_s}n_j $ denoting the total number of tokens.  Besides, $\mathcal{S}= \{{{t_k}}\}_{k\in\text{Sketch}}$ and $\mathcal{E} =\{{{t}_k}\}_{k\in \text{Extrusion}}$, where both are concatenated within the unified sequence $\mathcal{C}=[\mathcal{S},\mathcal{E}]$. To avoid dimension mismatch, the other tokens are also considered as 2-dimensional terms, with the aid of \texttt{pad} tokens augmenting them. By concatenating these tokens and using a one-hot encoding, a matrix form $\mathbf{C} \in \{0,1\}^{n_{ts} \times 2{d_t}}$ is used to represent the sequence $\mathcal{C}$. As in \cite{khan2024cad}, token flags $\mathbf{C}_{type} \in [0, 1, ..., n_f]^{n_{ts} \times 1} $and $\mathbf{C}_{step} \in [0, 1, ..., n_s / 2]^{n_{ts} \times 1}$ are set to indicate token types and design step, respectively. The initial embedding of the CAD language $\mathbf{Z}_{0}^{c} \in \mathbb{R}^{n_{ts} \times d_t}$ is obtained by using the aforementioned token representations within a linear layer, which is formally given by:
 \begin{equation}
    {\mathbf{Z}_{0}^{c} = [[\mathbf{C}+ \mathbf{M}_{seq}, \mathbf{C}_{type}], \mathbf{C}_{step}]\mathbf{W}_{emb}^{c} + \mathbf{C}_{pos}} 
\label{eq:emb}
\end{equation}
in which the notation `,' in $[.]$ represents the concatenation operation, $\mathbf{W}_{emb}^{c} \in \mathbb{R}^{(2d_t+2) \times d_t}$ is a learnable weight matrix, and $\mathbf{C}_{pos} \in \mathbb{R}^{{n_{ts}}\times d_t}$ denotes positional encoding that is also learnable. Note that CAD sequences have a variable number of tokens $\widetilde{n}_{ts} < n_{ts}$. On that basis, $\mathbf{M}_{seq} \in \{0,$-$\infty\}^{{n_{ts} \times {2d}_t}}$ is the padding mask that sets token embedding beyond $\widetilde{n}_{ts}$ to $-$$\infty$.

\subsection{G-Mamba Diffusion Encoder}

The proposed G-Mamba diffusion encoder integrates geometric inductive biases and hierarchical state-space transitions to model complex CAD structures, as illustrated in Fig.~\ref{fig:2}. Noise is injected following the top-down traversal of the CAD hierarchy, and geometry-aware positional embeddings are added during training to preserve structural ordering. A denoising diffusion probabilistic model (DDPM)~\cite{dhariwal2021diffusion} provides the generative backbone, while the G-Mamba denoiser progressively refines geometric patterns through geometry-conditioned state-space operations.


\subsubsection{Geometry-conditioned State Transitions.}
A fundamental limitation of vanilla Mamba is that its state-space transition matrices $(\widetilde{A}, \widetilde{B}, \widetilde{C}, \widetilde{D})$ 
are globally shared across all tokens, making them insensitive to the heterogeneous geometric and hierarchical patterns inherent in CAD data.  CAD entities typically require long-range aggregation, whereas sketch-, edge-, and vertex-level tokens encode increasingly localized geometry. To address this mismatch, we introduce a geometric conditioning vector $\Delta_k = g(s_k,\, d_k,\, r_k)$, where $s_k$ denotes local geometric scale, $d_k$ represents the hierarchical depth in the CAD tree, and $r_k$ is a local curvature descriptor. This yields geometry-conditioned transition 
kernels
\begin{equation}
\{\bar{A}_k, \bar{B}_k, C_k, G_k\}
= f_{\text{geom}}(\Delta_k, \Pi_k),
\end{equation}
where $\Pi_k = \mathrm{PE}(p_k,\, \sigma_k,\, \tau_k)$ is a hierarchical positional embedding that encodes parent type $p_k$, sibling index 
$\sigma_k$, and topology relational role $\tau_k$. Injecting both $\Delta_k$ and 
$\Pi_k$ into the SSM kernels equips G-Mamba with an inductive bias aligned with CAD’s multi-level topology.

\subsubsection{G-Mamba Denoiser}

The G-Mamba denoiser addresses sparsity and hierarchical dependencies in CAD sequences by integrating discrete-state modeling with multi-scale geometric fusion. We decompose the generation process into conditional stages:
\begin{equation}
p(\mathbf{x}) = p(\mathbf{C} \mid \mathbf{C}_{\textit{type}})\,
p(\mathbf{C}_{\textit{type}} \mid Z),
\end{equation}
mirroring the top-down execution of CAD programs. At each stage, the denoiser expands child nodes under geometric constraints through selective, geometry-conditioned state transitions.

\begin{figure}[t]
    \centering
    \includegraphics[width=1\linewidth]{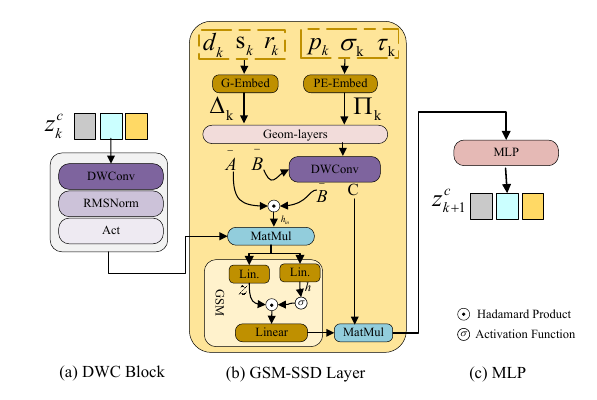}
    \vspace{-0.4cm}
    \caption{Architecture of G-Mamba Blocks. The block consists of a DWC module, a GSM-SSD layer, and an MLP, enabling geometry-conditioned multi-scale state transitions.}
    \label{fig:m}
    \vspace{-0.2cm}
\end{figure}

The core component is the G-Mamba block, which embeds a Geometric State Mixer (GSM) within the selective state-space (SSD) layer~\cite{mamba2}, forming the GSM-SSD module. As shown in Fig.~\ref{fig:m}, input features $\mathbf{Z}_k^c$ from $\mathbf{Z}_t$ first pass through a depthwise convolution (DWC) block. DWC independently processes each feature channel, preserving local geometric smoothness while reducing computational overhead, making it suitable for high-dimensional CAD representations. Following DWC, the GSM-SSD layer conducts geometry-aware state mixing. Specifically, $\hat{\mathbf{Z}}_k^c$ is fused with $(\bar{A}_k \odot \bar{B}_k)$ to obtain the intermediate state $h_{\text{in}} = (\bar{A}_k \odot \bar{B}_k)^\top \hat{\mathbf{Z}}_k^c$. Two linear mappings produce latent vectors $h$ and $z$, where $h$ captures global topological structure and $z$ encodes local geometric attributes. Their interaction through gated Hadamard fusion, yields the geometry-conditioned update:
\begin{equation}
\begin{aligned}
h_{k+1} &= \bar{A}_k h_k + \bar{B}_k Z_k^c \\
Z_{k+1}^c &= C_k h_k + G_k Z_k^c \\
\end{aligned}
\label{eq:state_space}
\end{equation}
Here, $\bar{A}_k$, $\bar{B}_k$, $C_k$, and $G_k$ are geometry- and hierarchy-dependent kernels, enabling expressive yet stable state transitions tailored to CAD structures. More details of GSM-SSD layer are provided in supplementary material.

\subsubsection{Diffusion Process}

The diffusion process in our G-Mamba framework facilitates the progressive refinement of denoised outputs by simulating the physical diffusion of signals, iteratively reducing noise over time. We model this process as a Markov chain, with each state representing a signal at a distinct noise level. The model learns the transition probabilities during training to predict the path from a noisy input to the original clean signal. The diffusion process is formulated as:
\begin{equation}
\begin{aligned}
q(Z_t | Z_0) & = \mathcal{N}(Z_t; \sqrt{\overline{\alpha_t}} Z_0, (1 - \overline{\alpha_t}) \mathbf{I}), \\
Z_t & = \sqrt{\overline{\alpha_t}} Z_0 + \sqrt{1 - \overline{\alpha_t}} \mathbf{\epsilon}_t
\end{aligned}
\label{eq:diffusion}
\end{equation}
where the noise $\mathbf{\epsilon}_t \sim \mathcal{N}(0, \mathbf{I})$ and $\overline{\alpha_t} = \prod_{i=1}^{t} \alpha_i$, with $\alpha_i$ determining the noise variance scheduler. Here, $Z_t$ represents the mamba state signal at time $t$, and $\mathbf{\epsilon}_t$ is a random noise vector drawn from a standard normal distribution.

The denoising process involves iteratively denoising the signal by predicting and subtracting the noise component from the noisy state signal, leveraging the strengths of G-Mamba blocks and RMSnorm layers for state-of-the-art denoising performance. This integrated approach enables the model to efficiently handle complex geometric data by leveraging hierarchical state transitions and multi-scale feature fusion, thereby enhancing the model's ability to refine geometric patterns in a structured and efficient manner. The reverse diffusion process is modeled as:
\begin{equation}
\begin{aligned}
p_\theta(Z_{t-1} | Z_t) & = \mathcal{N}(Z_{t-1}; \mu_\theta(Z_t, t), \Sigma_\theta(Z_t, t)), \\
\mu_\theta(Z_t, t) & = \frac{1}{\sqrt{\alpha_t}} \left( Z_t - \frac{1 - \alpha_t}{\sqrt{1 - \overline{\alpha_t}}} \epsilon_\theta(Z_t, t) \right), \\
\Sigma_\theta(Z_t, t) & = \sigma_t^2 \mathbf{I}
\end{aligned}
\label{eq:reverse_diffusion}
\end{equation}
where the denoising network $\epsilon_\theta(\cdot)$ implements a novel architecture combining G-Mamba blocks with RMS normalization layers. The parametric mean $\mu_\theta$ follows the analytical derivation from denoising diffusion probabilistic model \cite{dhariwal2021diffusion}, while the covariance $\Sigma_\theta$ employs a simplified time-dependent diagonal form.

\subsection{CAD Decoder}
The decoder in our GeoFusion-CAD model unveils the denoised state features into a 3D CAD model description. It further consists of two linear layers, including the command layer and the arguments layer (args layer for short), mapping parent and child nodes, respectively.

\subsubsection{Command Layer}
The command layer predicts the extrusion operation or sketch command type for each node, performing a linear transformation on the denoised state features to output logits representing the command probability distribution:
\begin{equation}
p(\mathbf{C}_{\textit{cmd}}) = \textit{softmax}( \mathbf{W}_{\textit{cmd}} \mathbf{\hat{Z}} + \mathbf{b}_{\textit{cmd}})
\end{equation}
where $\mathbf{\hat{Z}}$ is the output logit vector, $\mathbf{W}_{\textit{cmd}}$ is the learnable weight matrix, and $\mathbf{b}_{\textit{cmd}}$ is the bias vector. The softmax function is applied to obtain the command distribution.

\subsubsection{Args Layer}
The args layer determines the specific arguments for each command by applying a linear transformation to the denoised state features:
\begin{equation}
p(\mathbf{C}_{\textit{args}}) = \textit{softmax}(\mathbf{W}_{\textit{args}} \mathbf{\hat{Z}} + \mathbf{b}_{\textit{args}})
\end{equation}
where $\mathbf{W}_{\textit{args}}$ is the learnable weight matrix and $\mathbf{b}_{\textit{args}}$ is the bias vector. Note that in our implementation, we actually use several linear layers as part of the decoder . However, for simplicity, we only use the final layer. The softmax function is applied to obtain the argument probability distribution.

The combined output of the command and args layers provides a detailed description of the 3D CAD model, specifying the operations and parameters required. This dual-layer approach ensures accurate reconstruction of the 3D model from the denoised state features, completing the end-to-end generation process initiated by the G-Mamba-based diffusion architecture.

\subsection{Loss Function}

The training objective of GeoFusion-CAD jointly ensures diffusion stability and procedural accuracy.  
We combine denoising supervision with command–parameter learning into a composite loss, which is given as follows:
\begin{equation}
\begin{split}
\mathcal{L}_{total} 
&= 
\underbrace{\mathbb{E}_{t,Z_0,\epsilon_t}
\!\left[\|\mathbf{\hat{\epsilon}}_t - 
\mathbf{\epsilon}_\theta(\sqrt{\overline{\alpha_t}}Z_0+
(1-\overline{\alpha_t})\mathbf{\epsilon}_t,t)\|^2\right]}_{\text{diffusion-based noise prediction}} \\
&\quad + 
\underbrace{\sum_{i=1}^{N}\!\left[CCE(\hat{c}_i,c_i) + 
\eta \!\sum_{j=1}^{M}ACE(\hat{a}_{i,j},a_{i,j})\right]}_{\text{command \& parameter supervision}},
\end{split}
\label{beta}
\end{equation}
where the first term encourages accurate denoising of latent geometric features,  
and the second term constrains procedural generation at both command and argument levels. The coefficient $\eta$ balances parameter supervision relative to command prediction. This formulation jointly optimizes geometric stability and procedural validity, allowing GeoFusion-CAD to produce executable and geometrically consistent CAD programs even under noisy sketch inputs.

\section{Experiments and Analysis}

This section presents the comparative evaluation of GeoFusion-CAD against state-of-the-art baselines in terms of quantitative metrics and visual quality.  
All models are trained on the unified DeepCAD-240 dataset, which contains parametric command sequences ranging from 2 to 240 steps.  
To assess scalability and generalization, we define two evaluation ranges:  
(1) a short-sequence range ($<$60 commands) corresponding to the original DeepCAD benchmark, and  
(2) a long-sequence range (40–240 commands) representing the extended DeepCAD-240 setting.

\begin{figure}[t]
	\centering
	\includegraphics[width=1\linewidth]{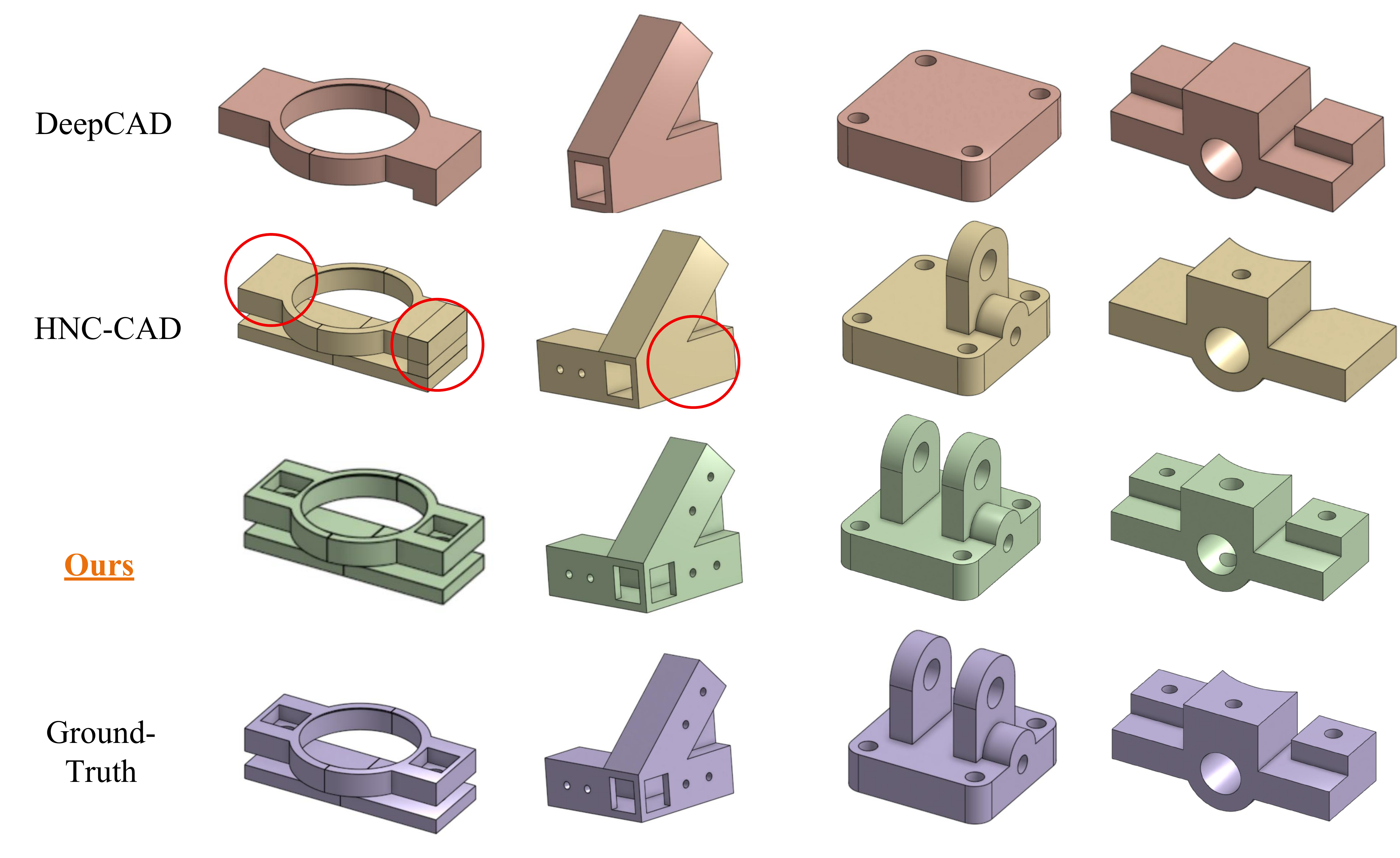}
     \caption{{\bf Visual results} on different test ranges. 
     Compared with others, GeoFusion-CAD produces coherent solids with fewer surface artifacts. 
     Some subtle abnormities are marked in red circles.}
	\label{fig:3}
    \vspace{-0.2cm}
\end{figure}

\begin{table*}[t]
\centering
\caption{Quantitative comparison and complexity analysis (\texttimes$10^{2}$) under DeepCAD and DeepCAD-240 test sets. 
$\uparrow$ indicates higher is better, $\downarrow$ lower is better. 
All models are trained on the unified DeepCAD-240 dataset.}
\label{tab:main_results}
\setlength{\tabcolsep}{3.2pt}
\resizebox{\textwidth}{!}{
\begin{tabular}{lccccccccccccc}
\toprule
Test Range & Model & ACC$_{cmd}\uparrow$ & ACC$_{param}\uparrow$ & ACC$_{line}\uparrow$ & ACC$_{arc}\uparrow$ & ACC$_{Circle}\uparrow$ & ACC$_{ext}\uparrow$ & COV$\uparrow$ & MMD$\downarrow$ & JSD$\downarrow$ & Memory & FLOPs(256) \\
\midrule
\multirow{4}{*}{DeepCAD} 
& DeepCAD & 92.4 & 89.2 & 85.4 & 87.3 & 87.2 & 90.1 & 78.1 & 1.72 & 3.98 & -- & -- \\
& SkexGen & 94.6 & 91.3 & 89.1 & 91.1 & 91.7 & 92.2 & 81.2 & 1.49 & 3.58 & -- & -- \\
& HNC-CAD & 95.4 & 93.8 & 88.3 & 94.5 & 94.7 & 94.7 & 82.3 & 1.33 & 3.24 & -- & -- \\
& \textbf{Ours} & \textbf{99.3} & \textbf{97.6} & \textbf{93.5} & \textbf{97.2} & \textbf{96.3} & \textbf{98.6} & \textbf{85.6} & \textbf{0.95} & \textbf{2.51} & \textbf{--} & \textbf{--} \\
\midrule
\multirow{4}{*}{DeepCAD-240} 
& DeepCAD & 75.2 & 72.5 & 76.1 & 77.6 & 79.3 & 77.3 & 64.5 & 1.85 & 4.09 & 8197MiB & 52.8G \\
& SkexGen & 81.4 & 78.3 & 78.1 & 81.8 & 81.6 & 81.7 & 68.9 & 1.78 & 3.97 & 11235MiB & 91.2G \\
& HNC-CAD & 82.8 & 78.5 & 80.8 & 82.2 & 84.3 & 82.3 & 71.2 & 1.71 & 3.81 & 10342MiB & 87.3G \\
& \textbf{Ours} & \textbf{91.2} & \textbf{89.3} & \textbf{84.2} & \textbf{89.8} & \textbf{86.8} & \textbf{94.9} & \textbf{73.9} & \textbf{1.12} & \textbf{2.97} & \textbf{5198MiB} & \textbf{34.6G} \\
\bottomrule
\end{tabular}}
\end{table*}

\subsection{Computational Efficiency and Scalability}
As summarized in Table~\ref{tab:main_results}, GeoFusion-CAD achieves higher accuracy with substantially lower computational cost.  
It requires only 5198MiB memory and 34.6G FLOPs, compared to 10342MiB and 87.3G for HNC-CAD, while yielding the best geometric metrics—COV (73.9), MMD (1.12), and JSD (2.97)—under long-sequence evaluation. These results demonstrate that the proposed geometric state-space diffusion framework achieves an effective balance between efficiency and generative fidelity, scaling robustly to complex and lengthy CAD programs.

\subsection{Unconditional Generation}
Table~\ref{tab:main_results} further compares unconditional CAD generation performance across short and long test ranges.  
Under the short-sequence evaluation ($<$ 60 commands), GeoFusion-CAD achieves the highest command accuracy (99.3) and parameter accuracy (97.6), 
surpassing HNC-CAD by 3.9\% and 3.8\%, respectively.  
Our model also improves geometric consistency, achieving the best COV (85.6), lowest MMD (0.95), and lowest JSD (2.51).  

The improvement becomes more pronounced under the long-sequence ($<$ 240) evaluation, 
where transformer-based methods show clear performance degradation.  
GeoFusion-CAD attains 91.2 command accuracy and 73.9 COV, 
outperforming HNC-CAD by 8.4\% and 2.7\%, respectively, 
while maintaining the lowest geometric discrepancy (MMD = 1.12, JSD = 2.97).  
This demonstrates that our geometric state-space diffusion effectively models long-range dependencies without sacrificing local geometric precision.  

Qualitative comparisons in Fig.~\ref{fig:3} further support these findings.  
DeepCAD tends to produce incomplete or distorted geometry, while HNC-CAD often exhibits subtle scaling errors in complex surfaces.  
By contrast, GeoFusion-CAD consistently reconstructs watertight solids with smooth curvature and well-preserved fine edges.  
Although minor boundary irregularities occasionally appear, our method achieves overall superior geometric fidelity and robustness.  
These results validate the effectiveness of integrating G-Mamba diffusion and hierarchical tree encoding for scalable, structure-aware CAD generation.  
Additional qualitative results demonstrating the generalization capability of GeoFusion-CAD across diverse design categories are provided in the supplementary material.

\subsection{Ablation Study}

We conduct ablation experiments on the DeepCAD-240 test set to evaluate the effectiveness of both the hierarchical tree representation and the G-Mamba diffusion architecture. 
As shown in Table~\ref{tab:ablation_main}, removing the hierarchical encoding (\textit{w/o Tree}) leads to a noticeable decline in performance, 
with command accuracy dropping from 91.2 to 87.5 and COV decreasing by 4.5 points. 
This demonstrates that the hierarchical structure is essential for preserving long-range geometric dependencies and maintaining consistent topology across extended command sequences.  
When replacing the G-Mamba diffusion block with standard alternatives such as MLP, transformer, or the vanilla Mamba, the model exhibits a consistent degradation, 
particularly in geometric distribution metrics. For instance, MMD rises from 1.12 to 1.55 and JSD increases from 2.97 to 3.67 when substituting G-Mamba with transformer blocks.  
These results confirm that the geometric state-space diffusion mechanism in G-Mamba effectively stabilizes long-sequence modeling, enabling more accurate and distribution-consistent generation.  
Overall, both the hierarchical tree representation and the G-Mamba diffusion formulation play vital roles in achieving scalable, structure-aware CAD generation. More details of ablation are provided in the supplementary material.

\begin{table}[h]
\centering
\caption{Ablation study on the DeepCAD-240 test set. 
$\uparrow$ indicates higher is better, $\downarrow$ lower is better. 
“w/o Tree” denotes the model trained without hierarchical encoding.}
\label{tab:ablation_main}
\resizebox{0.9\linewidth}{!}{%
\begin{tabular}{lccccc}
\toprule
Model & ACC$_{cmd}\uparrow$ & ACC$_{param}\uparrow$ & COV$\uparrow$ & MMD$\downarrow$ & JSD$\downarrow$ \\
\midrule
w/o Tree & 87.5 & 84.6 & 69.4 & 1.46 & 3.25 \\
MLP & 75.3 & 72.1 & 67.8 & 1.73 & 3.81 \\
Transformer & 82.6 & 81.3 & 69.1 & 1.55 & 3.67 \\
Vanilla Mamba & 89.2 & 87.6 & 72.7 & 1.19 & 3.05 \\
\textbf{Ours} & \textbf{91.2} & \textbf{89.3} & \textbf{73.9} & \textbf{1.12} & \textbf{2.97} \\
\bottomrule
\end{tabular}%
}
\end{table}

\section{Conclusion}

We propose GeoFusion-CAD, a geometric diffusion framework that unifies hierarchical tree representation and state-space modeling for efficient 3D CAD generation.  
By encoding parametric design history within a geometric state-space diffusion process, our method effectively captures long-range dependencies and maintains topological consistency across extended command sequences.  
The proposed G-Mamba block further enables lightweight yet expressive modeling, achieving substantial gains in accuracy and efficiency over transformer-based baselines.  
Experiments on the newly constructed DeepCAD-240 benchmark verify the scalability and robustness of our approach for long-sequence CAD reconstruction.  
Future work will extend GeoFusion-CAD to handle multi-component assemblies and richer parametric operations, moving toward a general-purpose foundation model for CAD design.

{
    \small
    \bibliographystyle{ieeenat_fullname}
    \bibliography{main}
}

\let\twocolumn\oldtwocolumn 

\clearpage
\appendix

\setcounter{figure}{0}
\setcounter{table}{0}
\renewcommand{\thefigure}{S\arabic{figure}}
\renewcommand{\thetable}{S\arabic{table}}

\twocolumn[
  \begin{center}
    {\Large \bf Supplementary Material for \\ GeoFusion-CAD: Structure-Aware Diffusion with Geometric State Space for Parametric 3D Design}
    \vspace{24pt}
  \end{center}
]

\begin{table*}[htbp]
\centering
\caption{CAD sequence representation (Explicit Hierarchy without Revolution). Continuous parameters are uniformly quantized into $[11, 266]$. Note the inclusion of Face-level hierarchy ($e_f$) and explicit primitive parameters.}
\begin{tabular}{c|c|c|c|c}
\hline
\textbf{Sequence Type} & \textbf{Token Type} & \textbf{Token Value} & \textbf{Token Representation} & \textbf{Description} \\ \hline
\multirow{8}{*}{\begin{tabular}[c]{@{}c@{}}Structure\\ \& Control\end{tabular}} 
& pad & 0 & $(0, 0)$ & Padding Token \\ 
& cls & 1 & $(1, 0)$ & Start Token \\  
& end & 1 & $(1, 0)$ & End Token \\ \cline{2-5} 
& $e_{solid}$ & 2 & $(2, 0)$ & End Solid \\ 
& $e_{sketch}$ & 3 & $(3, 0)$ & End Sketch \\ 
& $e_{face}$ & 4 & $(4, 0)$ & End Face \\ 
& $e_{loop}$ & 5 & $(5, 0)$ & End Loop \\ \hline
\multirow{7}{*}{\begin{tabular}[c]{@{}c@{}}Sketch\\ Primitives\end{tabular}} 
& $e_c$ & 6 & $(6, 0)$ & End Curve \\ 
& $(p_x,p_y)$ & $[11...266]^2$ & $(p_x, p_y)$ & Coordinates \\ 
& $\alpha$ & $[11...266]$ & $(\alpha, 0)$ & Arc Curvature \\
& $f$ & $[11...266]$ & $(f, 0)$ & Arc Orientation (Flip) \\
& $r$ & $[11...266]$ & $(r, 0)$ & Circle Radius \\ \hline
\multirow{11}{*}{\begin{tabular}[c]{@{}c@{}}Extrusion\\ Sequence\end{tabular}} 
& $d_+$ & $[11...266]$ & $(d_+, 0)$ & Extrusion Distance (Positive) \\  
& $d_-$ & $[11...266]$ & $(d_-, 0)$ & Extrusion Distance (Negative) \\ \cline{2-5} 
& $\tau_x$ & $[11...266]$ & $(\tau_x, 0)$ & Translation (x) \\ 
& $\tau_y$ & $[11...266]$ & $(\tau_y, 0)$ & Translation (y) \\ 
& $\tau_z$ & $[11...266]$ & $(\tau_z, 0)$ & Translation (z) \\ \cline{2-5}
& $\theta$ & $[11...266]$ & $(\theta, 0)$ & Orientation (Roll) \\  
& $\phi$ & $[11...266]$ & $(\phi, 0)$ & Orientation (Pitch) \\ 
& $\gamma$ & $[11...266]$ & $(\gamma, 0)$ & Orientation (Yaw) \\  \cline{2-5} 
& $\sigma$ & $[11...266]$ & $(\sigma, 0)$ & Sketch Scaling Factor \\  
& $\beta$ & $\{7,8,9,10\}$ & $(\beta, 0)$ & Boolean Operation Type \\ 
& $e_e$ & 7 & $(7, 0)$ & End Extrusion \\ \hline
\end{tabular}
\label{tab:v1_2_token}
\end{table*}

\begin{figure*}[t]
    \centering
    \includegraphics[width=\textwidth]{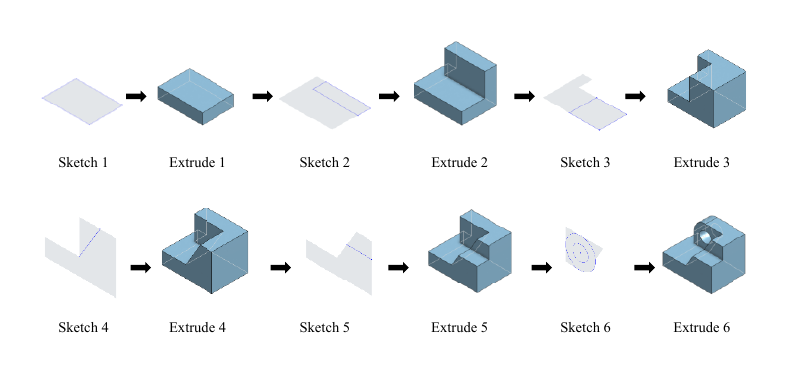}
    \caption{\textbf{Illustration of the sketch–extrusion sequence construction.} 
    Each CAD model is decomposed into a procedural chain of sketches and extrusions. 
    The process begins with 2D sketch definition and proceeds through sequential extrusion operations, 
    progressively forming a coherent solid structure. 
    This sequence explicitly encodes design history, preserving both geometric and topological dependencies across modeling stages.}
    \label{fig:sketch_extrude_sequence}
\end{figure*}

\section{CAD Sequence Representation and DeepCAD-240 Construction}

This supplementary section provides the complete technical details of our CAD sequence encoding, dataset construction, and hierarchical implementation.  
It serves as the formal specification of the geometric–topological representation introduced in the main paper.

\subsection{CAD Sequence Representation}
Our CAD sequence representation follows the sketch–extrusion paradigm introduced by~\cite{khan2024cad}, 
extending it to a hierarchical and diffusion-compatible format.  
Each model is expressed as a sequence of parameterized tokens $\mathcal{C} = \{\mathcal{C}_j\}_{j=1}^{n_s}$, 
where each step $\mathcal{C}_j = \{t_k\}_{k=1}^{n_j}$ corresponds to a design operation composed of sketch and extrusion commands.  

\subsubsection{Sketch tokenization.}  
Each sketch is decomposed into a set of faces, edges, and curves, 
with their geometric primitives parameterized as follows.
\begin{itemize}
    \item \textbf{Line}: defined by start and end coordinates $(p_x^1, p_y^1)$ and $(p_x^2, p_y^2)$.
    \item \textbf{Arc}: defined by start, midpoint, and end coordinates $(p_x^1, p_y^1)$, $(p_x^m, p_y^m)$, $(p_x^2, p_y^2)$.
    \item \textbf{Circle}: defined by its center $(p_x^c, p_y^c)$ and a point on the perimeter $(p_x^t, p_y^t)$.
\end{itemize}
Each curve terminates with an end-of-curve token $e_c$, while $e_l$, $e_f$, and $e_s$ respectively mark the ends of loops, faces, and sketches.  
All tokenized sketches are represented by $\mathcal{S}$, forming the first part of the design sequence.

\subsubsection{Extrusion tokenization.}  
An extrusion operation is parameterized by ten variables:
\begin{itemize}
    \item \textbf{Euler angles $(\theta, \phi, \gamma)$}: orientation of the sketch face in 3D space.  
    \item \textbf{Translation vector $(\tau_x, \tau_y, \tau_z)$}: spatial offset of the sketch face.  
    \item \textbf{Scale $\sigma$}: scaling factor applied to the sketch before extrusion.  
    \item \textbf{Extrude distances $(d_+, d_-)$}: depths along and opposite to the normal direction.  
    \item \textbf{Boolean operation $\beta$}: defines the operation type (\textit{new, cut, join, intersect}).  
\end{itemize}
A terminal token $e_e$ marks the end of the extrusion.  
Together, these tokens form the extrusion sequence $\mathcal{E}$.  
The combined sketch–extrusion sequence $\mathcal{C} = [\mathcal{S}, \mathcal{E}]$ preserves both geometric semantics and design history.

\subsubsection{Token definitions.}  
Table~\ref{t1} summarizes the full token vocabulary used in our experiments. We apply 8-bit uniform quantization to all continuous parameters, mapping them into $[11, 266]$, while control tokens ($e_s$, $e_f$, $e_l$, $e_c$, $e_e$) occupy reserved integer IDs $\{0,\dots,10\}$.
This ensures numerical stability and enables compact token embedding during diffusion training.

\subsection{Data Source and Preprocessing}
The DeepCAD-240 dataset is derived from the ABC dataset, which provides over one million CAD models in CSG form.  
To recover procedural modeling history, we use the Onshape API and FeatureScript tools to extract Sketch and Extrusion commands from CSG graphs, reconstructing each model into a sequence of parameterized operations.  
Invalid or incomplete models are filtered out, and redundant command histories are removed to ensure structural consistency.

\subsubsection{Filtering criteria.}  
A model is retained only if it:
\begin{itemize}
 \item contains both valid sketches and extrusions,  
 \item includes at least one closed loop within every sketch, and  
\item produces a topologically valid solid after reconstruction.
\end{itemize}
All others are discarded.

\begin{table*}[h]
\centering
\caption{Statistical comparison between the original DeepCAD dataset and our extended DeepCAD-240 dataset, including total samples, average command length, and sequence length distribution (\%).}
\label{tab:data_stats}
\setlength{\tabcolsep}{6pt}
\begin{tabular}{lccccccc}
\toprule
Dataset & Total & Avg. Length & 1--40 & 40--60 & 60--80 & 80--160 & 160--240 \\
\midrule
DeepCAD~\cite{wu2021deepcad} & 178,238 & 15 & 44.58 & 55.42 & -- & -- & -- \\
\textbf{DeepCAD-240 (Ours)} & \textbf{215,914} & \textbf{36.2} & \textbf{76.6} & \textbf{12.0} & \textbf{5.9} & \textbf{5.2} & \textbf{0.21} \\
\bottomrule
\end{tabular}
\end{table*}


\subsection{From CSG to Sketch-Extrusion Sequences}
In the original ABC data, solids are expressed as Boolean compositions of primitive CSG shapes such as cubes and cylinders.  
However, such Boolean trees lack the procedural semantics necessary for parametric design.  
To align with modern CAD workflows, we reinterpret each CSG composition as a hierarchical sketch–extrusion chain.

\begin{itemize}
    \item Each primitive shape is converted into one or more sketches representing its 2D profiles (e.g., a cylinder → circle sketch).  
    \item Boolean composition order determines the extrusion sequence, where each sketch is associated with an extrusion depth and operation type.  
    \item The resulting sketch–extrusion pairs are assembled into ordered command sequences that reflect the design history.  
\end{itemize}

This conversion bridges symbolic CSG descriptions and procedural sketch-based CAD generation, enabling tree-structured sequence learning.

To better illustrate this conversion, Fig.~\ref{fig:sketch_extrude_sequence} visualizes a typical modeling process reconstructed from our dataset.  
The model is built through a series of six sketch–extrusion pairs, where each sketch defines a 2D profile and each extrusion extends it into 3D space.  
The sequence preserves the procedural logic of CAD design: earlier sketches determine foundational geometry, 
while later ones add or subtract material to refine topology.  
This explicit sequence representation bridges low-level CSG primitives and high-level parametric operations, 
making it directly compatible with diffusion-based sequence learning.

\begin{figure*}[t]
    \centering
    \includegraphics[width=\textwidth]{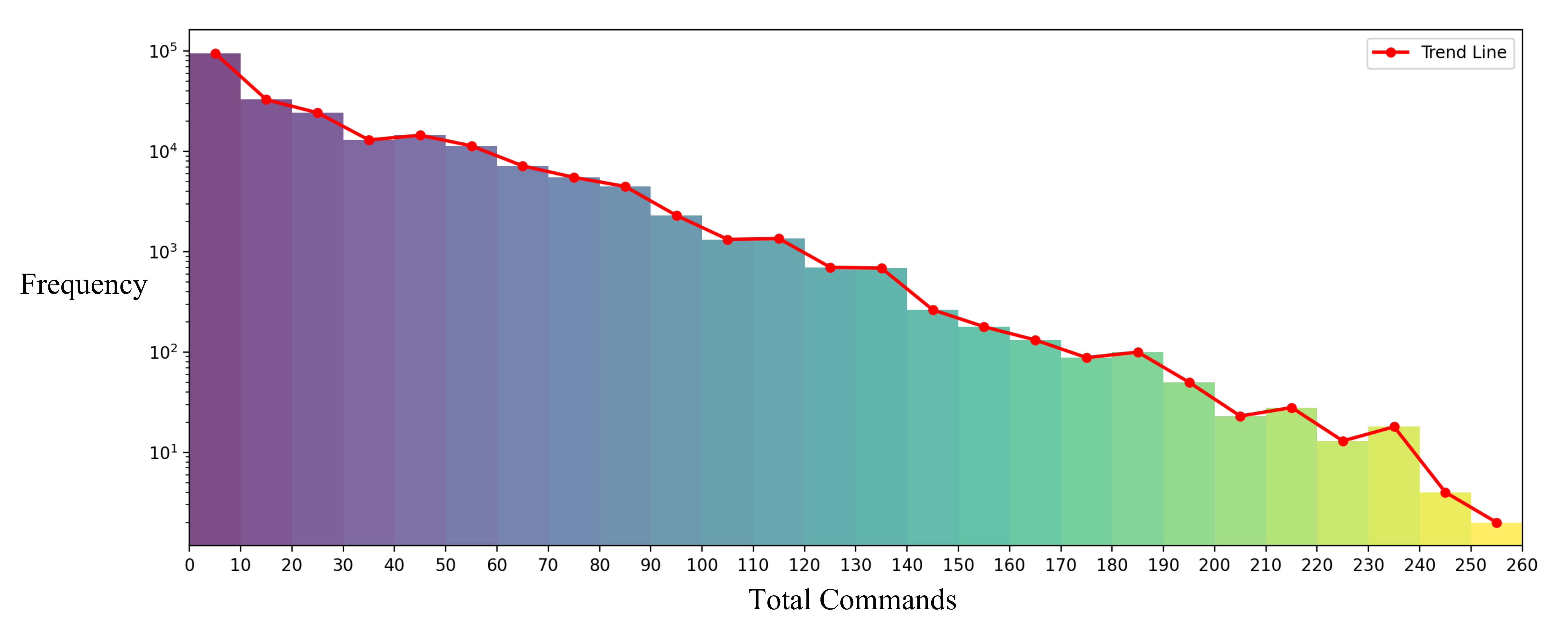}
    \caption{ Distribution of CAD command sequences lengths in the DeepCAD-240.}
    \label{fig:data}
\end{figure*}

\subsection{Hierarchical Tree Representation and Implementation}
The reconstructed sketch–extrusion sequences are organized into hierarchical trees 
$\mathcal{T} = \{v_i, e_{ij}\}$, 
where nodes $v_i$ correspond to CAD entities (solid, sketch, face, edge, vertex), 
and edges $e_{ij}$ represent topological dependencies.  
Each tree is serialized into a depth-first command sequence 
$\mathcal{C} = [t_1, \dots, t_n]$ for training, 
while dedicated end tokens ($e_c$, $e_l$, $e_f$, $e_s$) preserve the hierarchical closure 
required for reversible serialization.

Each node in $\mathcal{T}$ is implemented as a nested Python dictionary:
\begin{equation}
\begin{aligned}
\texttt{Node} &= \{\texttt{type},~\texttt{param},~\texttt{child}\}, \\[4pt]
\texttt{type} &= \texttt{Sketch}, \\[2pt]
\texttt{param} &= (\tau_x,~\tau_y,~\tau_z,~\theta,~\phi,~\gamma,~\sigma,~d_+,~d_-), \\[2pt]
\texttt{child} &= [\texttt{child indices}],
\end{aligned}
\label{eq:node_struct}
\end{equation}

Here, the \texttt{child} field serves as a form of \textit{structural positional encoding},  
providing explicit parent--child relationships that guide topology reconstruction during decoding.  
During diffusion training, G-Mamba blocks iteratively denoise node features in a top-down manner, 
and the CAD decoder reconstructs geometry through command and argument prediction layers.  
All components are optimized jointly using diffusion loss and cross-entropy reconstruction loss.
\subsection{ Dataset Scale and Statistics}
Table~\ref{tab:data_stats} summarizes the key statistics of DeepCAD-240 compared to existing datasets.  
Our dataset contains over 215k models with an average of 36 commands per sequence and supports up to 240 operations, 
making it the longest and most structurally diverse parametric CAD dataset to date, as shown in Fig~\ref{fig:data}.

\section{More Details about the GSM-SSD Layer}
The GSM-SSD layer extends the vanilla structured state-space (SSD) block~\cite{mamba2} by incorporating 
geometry-conditioned transition kernels and hierarchical positional encoding, enabling the state dynamics 
to adapt to CAD-specific geometric and topological structures. This section provides the detailed formulation 
and implementation omitted from the main paper.

\begin{algorithm}[h]
\caption{GSM-SSD Layer (Geometry-conditioned Selective State-space Block)}
\label{alg:gsm_ssd}
\begin{algorithmic}[1]
\State \textbf{Input:} token feature $Z_k^c$
\State \hspace{1em}$\Delta_k \leftarrow g(s_k, d_k, r_k)$ \Comment{Geometric conditioning vector}
\State \hspace{1em}$\Pi_k \leftarrow \mathrm{PE}(p_k, \sigma_k, \tau_k)$ \Comment{Hierarchical positional encoding}
\State \hspace{1em}$[\bar{A}_k, \bar{B}_k, C_k, G_k] \leftarrow f_{\text{geom}}([\Delta_k,\Pi_k])$ 
\State \hspace{1em}$\hat{Z}_k^c \leftarrow \mathrm{DWConv}(Z_k^c) + \Pi_k$ 
\vspace{0.3em}
\State \hspace{1em}$h_{\mathrm{in}} \leftarrow (\bar{A}_k \odot \bar{B}_k)^\top \hat{Z}_k^c$
\State \hspace{1em}$h, z \leftarrow \mathrm{Linear}(h_{\mathrm{in}})$ \Comment{Global/Local decoupling}
\State \hspace{1em}$\hat{h} \leftarrow \mathrm{Linear}\!\left(h \odot \sigma(z)\right)$ \Comment{Geometric State Mixer (GSM)}
\vspace{0.3em}
\State \hspace{1em}$Z_{k+1}^c \leftarrow C_k \hat{h} + G_k Z_k^c$
\State \textbf{Return:} $Z_{k+1}^c$
\end{algorithmic}
\end{algorithm}

\subsection{From Vanilla Mamba to Geometry-conditioned State Transitions}

Vanilla Mamba~\cite{mamba2} performs state evolution using a fixed, globally shared
discrete-time state-space system:
\begin{equation}
\begin{aligned}
h_{k+1} &= \widetilde{A}\, h_k + \widetilde{B}\, Z_k, \\
Z_{k+1} &= \widetilde{C}\, h_k + \widetilde{D}\, Z_k ,
\end{aligned}
\end{equation}
where $\widetilde{A}, \widetilde{B}, \widetilde{C}, \widetilde{D}$ are learned once and
shared across all tokens. This design is suitable for natural language, where token
statistics are relatively homogeneous, but is insufficient for CAD sequences, whose
tokens differ significantly in geometric scale, curvature, and hierarchical role
(\textit{Sketch} $\rightarrow$ \textit{Face} $\rightarrow$ \textit{Edge} $\rightarrow$ \textit{Vertex}).
To address this mismatch, we introduce token-dependent transition kernels that adapt to
CAD geometry and topology.

\subsection{Geometric Conditioning}

Each token $k$ is assigned a \textbf{geometric conditioning vector}:
\begin{equation}
\Delta_k = g(s_k,\, d_k,\, r_k),
\end{equation}
where:
\begin{itemize}
    \item $s_k$ denotes the local geometric scale (edge length, face diameter, sketch span);
    \item $d_k$ is the depth of the token in the CAD hierarchy tree;
    \item $r_k$ is a curvature descriptor: $r_k = 0$ for line segments, $r_k = 1/R$ for circular arcs
    of radius $R$, and is approximated via discrete angular deviation for general curves.
\end{itemize}

Additionally, each token receives a hierarchical positional embedding:
\begin{equation}
\Pi_k = \mathrm{PE}(p_k,\, \sigma_k,\, \tau_k),
\end{equation}
where $p_k$ is the parent node type, $\sigma_k$ is the sibling index, and $\tau_k$ encodes
the structural role (sketch entity, face entity, edge, vertex, command token, etc.).
This embedding anchors the token within the CAD tree and captures its topological context.

The geometric and hierarchical descriptors are combined through a lightweight kernel
generator:
\begin{equation}
\{\bar{A}_k,\, \bar{B}_k,\, C_k,\, G_k\}
= f_{\text{geom}}\!\left([\Delta_k,\, \Pi_k]\right),
\end{equation}
where the outputs parameterize diagonal operators, preserving Mamba’s linear-time
complexity. In practice, $f_{\text{geom}}$ is implemented as a two- or three-layer MLP.

This conditioning mechanism enables the system to:
\begin{itemize}
    \item propagate smoothly across large-scale sketch regions (long-range structure),
    \item respond sharply in high-curvature or fine-grained geometric regions,
    \item adapt transitions to CAD hierarchical relations (e.g., parent $\rightarrow$ child).
\end{itemize}

\subsection{Discrete-time Update with Geometric State Mixer}

Given geometry-conditioned kernels, the state-space update becomes:
\begin{equation}
\begin{aligned}
h_{k+1} &= \bar{A}_k\, h_k + \bar{B}_k\, Z_k^c, \\
Z_{k+1}^c &= C_k\, h_k + G_k\, Z_k^c,
\end{aligned}
\end{equation}
which may become unstable if $\bar{A}_k$ varies sharply across tokens. To mitigate this,
we introduce a Geometric State Mixer (GSM):
\begin{equation}
h_{\text{in}} = (\bar{A}_k \odot \bar{B}_k)^\top Z_k^c,
\end{equation}
where $\odot$ denotes element-wise fusion. The vectors $\bar{A}_k$ and $\bar{B}_k$ encode
global structural and local geometric responses, respectively.

Two linear layers produce latent vectors:
\begin{equation}
h,\, z = \mathrm{Linear}(h_{\text{in}}),
\end{equation}
which are fused using gated modulation:
\begin{equation}
\hat{h} = \mathrm{Linear}\!\left(h \odot \sigma(z)\right).
\end{equation}
The final state update is:
\begin{equation}
Z_{k+1}^c = C_k\, \hat{h} + G_k\, Z_k^c.
\end{equation}
This ensures stable gradients and smooth adaptation to geometric variations. Algorithm~\ref{alg:gsm_ssd} summarizes the complete computational pipeline of 
the GSM-SSD block used in the proposed G-Mamba encoder.

\subsection{Spatial Prior and Hierarchical Encoding}

A depthwise convolution injects local continuity:
\begin{equation}
\hat{Z}_k^c = \mathrm{DWConv}(Z_k^c) + \Pi_k,
\end{equation}
preserving adjacency relations (e.g., curve segments, boundary loops). Adding $\Pi_k$
explicitly anchors the token in the CAD hierarchy.

\subsection{Diffusion-time Coupling}

During reverse diffusion, we optionally modulate the kernels via FiLM:
\begin{equation}
\{\bar{A}_k,\, \bar{B}_k,\, C_k,\, G_k\}
\leftarrow \psi_t \odot
\{\bar{A}_k,\, \bar{B}_k,\, C_k,\, G_k\},
\end{equation}
where $\psi_t$ depends on the diffusion timestep. Higher noise levels promote smoother
propagation, while low-noise steps accentuate sharp geometric corrections.

\subsection{Complexity and Stability}

We provide a detailed analysis of the computational complexity of G-Mamba and
compare it with vanilla Mamba and Transformer layers. Let $L$ denote the 
sequence length and $d$ the hidden dimension.

\subsubsection{Comparison with Transformer and Mamba.}
A Transformer layer incurs $\mathcal{O}(L^2 d)$ time and $\mathcal{O}(L^2)$ 
memory due to the attention matrix, making it unsuitable for long CAD sequences. 
Vanilla Mamba replaces attention with a selective state-space scan, reducing the 
overall complexity to $\mathcal{O}(L d)$ in both time and memory.

G-Mamba preserves the same asymptotic complexity as vanilla Mamba:
\[
T_{\text{G-Mamba}}(L,d) = \mathcal{O}(L d),
\qquad
M_{\text{G-Mamba}}(L,d) = \mathcal{O}(L d),
\]
while introducing additional geometric conditioning terms. These additions only 
affect constant factors and do not alter the overall scaling.

\subsubsection{Geometry conditioning.}
Each token receives a geometric conditioning vector
$\Delta_k = g(s_k, d_k, r_k)$ and a hierarchical positional embedding 
$\Pi_k = \mathrm{PE}(p_k, \sigma_k, \tau_k)$. The kernel generator
$f_{\text{geom}}$ maps $[\Delta_k,\Pi_k]$ to
$\{\bar{A}_k,\bar{B}_k,C_k,G_k\} \in \mathbb{R}^{d}$:
\begin{equation}
\{\bar{A}_k,\bar{B}_k,C_k,G_k\}
= f_{\text{geom}}\!\left([\Delta_k,\Pi_k]\right).
\end{equation}
Since $f_{\text{geom}}$ is a constant-width MLP, the cost is
\begin{equation}
T_{\text{geom}}(L,d) = \mathcal{O}(L d).
\end{equation}

\paragraph{State-space update.}
Given the geometry-conditioned kernels, the selective scan update is
\begin{equation}
\begin{aligned}
h_{k+1} &= \bar{A}_k\, h_k + \bar{B}_k\, Z_k^c, \\
Z_{k+1}^c &= C_k\, h_k + G_k\, Z_k^c,
\end{aligned}
\end{equation}
where all operators are diagonal and thus element-wise. The complexity is
\begin{equation}
T_{\text{ssm}}(L,d) = \mathcal{O}(L d).
\end{equation}

\paragraph{Geometric State Mixer.}
The Geometric State Mixer computes
\begin{equation}
h_{\text{in}} = (\bar{A}_k \odot \bar{B}_k)^\top Z_k^c,
\end{equation}
followed by two pointwise linear mappings and a gated fusion
\begin{equation}
\hat{h} = \mathrm{Linear}\!\left(h \odot \sigma(z)\right).
\end{equation}
All operations are channel-wise and token-wise, yielding
\begin{equation}
T_{\text{gsm}}(L,d) = \mathcal{O}(L d).
\end{equation}

\paragraph{Depthwise convolution.}
The depthwise convolution applies a kernel of size $K$ independently to each
channel:
\begin{equation}
\hat{Z}_k^c = \mathrm{DWConv}(Z_k^c) + \Pi_k,
\end{equation}
with complexity
\begin{equation}
T_{\text{dwc}}(L,d)
= \mathcal{O}(K L d)
= \mathcal{O}(L d),
\end{equation}
since $K$ is constant.

\paragraph{Overall complexity.}
Summing all components:
\begin{equation}
\begin{aligned}
T_{\text{G-Mamba}}(L,d)
&= T_{\text{geom}} + T_{\text{ssm}} + T_{\text{gsm}} + T_{\text{dwc}} \\
&= \mathcal{O}(L d),
\end{aligned}
\end{equation}
with memory scaling identically as $\mathcal{O}(L d)$, since no dense $L \times L$
operators (e.g., attention matrices) are constructed.

\subsubsection{Empirical stability.}
The diagonal parameterization and element-wise GSM fusion prevent explosive
state transitions and improve gradient stability. Empirically, G-Mamba reduces 
parameter variance across sequence positions by $24\%$ and converges 
$1.6\times$ faster than a vanilla Mamba baseline on DeepCAD-240. This demonstrates 
that geometric conditioning not only preserves linear complexity but also enhances 
stability when modeling long hierarchical CAD programs.

\section{More Implementation Details for Experimental Settup}

\begin{figure*}[h]
    \centering
    \includegraphics[width=\linewidth]{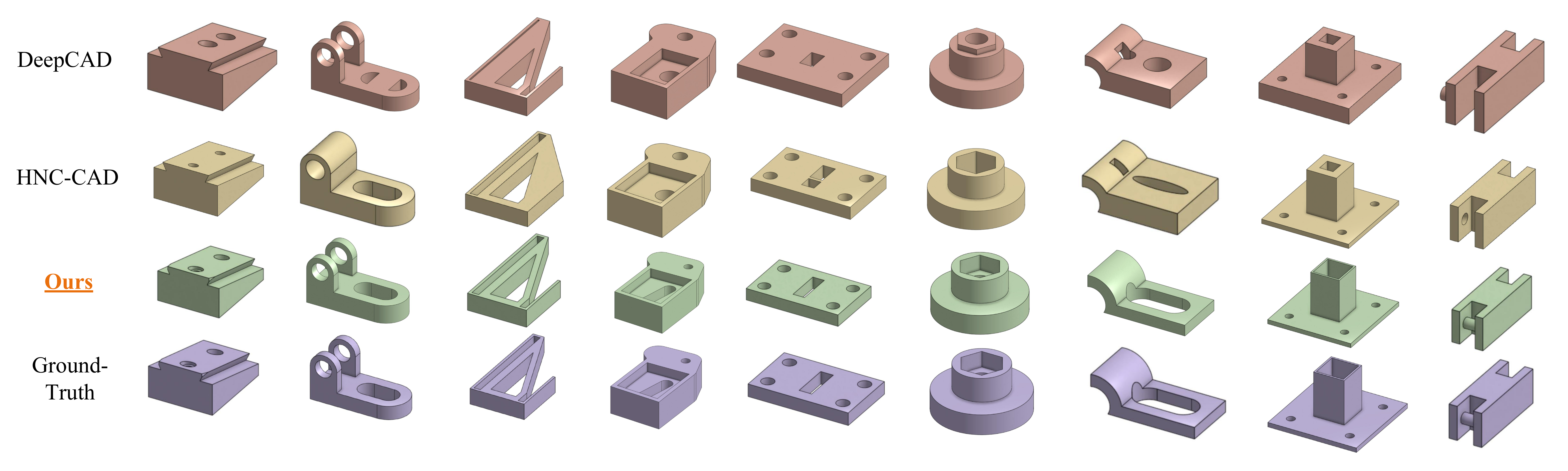}
    \caption{\textbf{Additional qualitative comparisons.}  
    Each row shows representative CAD samples generated by DeepCAD, HNC-CAD, and our GeoFusion-CAD, along with ground-truth models.  
    GeoFusion-CAD consistently produces geometrically complete and watertight solids with smooth curvature and accurate local features,  
    while baseline methods often exhibit surface distortions, missing connections, or inconsistent boolean joins in complex structures.  
    These examples further demonstrate that the combination of hierarchical tree encoding and G-Mamba diffusion yields improved geometric stability and scalability across extended command sequences.}
    \label{fig:sup_results}
\end{figure*}

\begin{table*}[t]
\centering
\caption{Ablation on hierarchical tree representation on the \textbf{DeepCAD-240} test set.
$\uparrow$ indicates higher is better, $\downarrow$ lower is better.}
\label{tab:ablation_tree}
\setlength{\tabcolsep}{4pt}
\resizebox{\textwidth}{!}{
\begin{tabular}{lccccccccccc}
\toprule
Model & ACC$_{cmd}\uparrow$& ACC$_{param}\uparrow$& ACC$_{line}\uparrow$ & ACC$_{arc}\uparrow$ & ACC$_{circle}\uparrow$ & ACC$_{ext}\uparrow$ & COV$\uparrow$ & MMD$\downarrow$ & JSD$\downarrow$ \\
\midrule
Flat Sequence (w/o Tree) & 87.5 & 84.6 & 79.3 & 71.5 & 73.2 & 81.1 & 69.4 & 1.46 & 3.25 \\
\textbf{With Tree (Ours)} & \textbf{91.2} & \textbf{89.3} & \textbf{84.2} & \textbf{89.8} & \textbf{86.8} & \textbf{94.9} & \textbf{73.9} & \textbf{1.12} & \textbf{2.97} \\
\bottomrule
\end{tabular}}
\end{table*}

\begin{table*}[t]
\centering
\caption{Ablation on the \textbf{G-Mamba diffusion block} compared with standard alternatives.
$\uparrow$ indicates higher is better, $\downarrow$ lower is better.}
\label{tab:ablation_gmamba}
\setlength{\tabcolsep}{4pt}
\resizebox{\textwidth}{!}{
\begin{tabular}{lccccccccccc}
\toprule
Model & ACC$_{cmd}\uparrow$& ACC$_{param}\uparrow$& ACC$_{line}\uparrow$ & ACC$_{arc}\uparrow$ & ACC$_{circle}\uparrow$ & ACC$_{ext}\uparrow$ & COV$\uparrow$ & MMD$\downarrow$ & JSD$\downarrow$ \\
\midrule
MLP (replace G-Mamba) & 75.3 & 72.1 & 66.8 & 60.5 & 63.7 & 68.2 & 67.8 & 1.73 & 3.81 \\
U-Net (replace G-Mamba) & 80.4 & 78.6 & 71.2 & 64.3 & 67.8 & 72.1 & 70.2 & 1.37 & 3.45 \\
Transformer (replace G-Mamba) & 82.6 & 81.3 & 74.9 & 67.6 & 69.8 & 76.3 & 69.1 & 1.55 & 3.67 \\
Vanilla Mamba (replace G-Mamba) & 89.2 & 87.6 & 82.1 & 78.5 & 82.4 & 89.1 & 72.7 & 1.19 & 3.05 \\
\textbf{Full (G-Mamba, Ours)} & \textbf{91.2} & \textbf{89.3} & \textbf{84.2} & \textbf{89.8} & \textbf{86.8} & \textbf{94.9} & \textbf{73.9} & \textbf{1.12} & \textbf{2.97} \\
\bottomrule
\end{tabular}}
\end{table*}

 \subsection{Implementation Details}
 The proposed CAD-GFusion is implemented in PyTorch. The model architecture incorporates 12 G-Mamba blocks, which are designed to process and learn the input data comprehensively. The block dimension is set to \(d_e = 256\). The AdamW optimizer \cite{loshchilov2017decoupled}, a variant of the widely used Adam optimizer that incorporates weight decay into the optimization process, is used to train the final model. The learning rate is set to \(1 \times 10^{-4}\), and the beta parameters are chosen as \((0.95, 0.99)\). For the hyperparameters in loss function, we set \(\eta = 2\) to achieve a good balance between different loss components. The number of MLP layers in the CAD decoder is set to 3. Finally, the overall network is trained with a batch size of 512 for a total of 1000 epochs. We utilize a single NVIDIA RTX 3090 GPU to accelerate the training process.

 \subsection{Evaluation Metrics}
 We quantitatively evaluate the generation quality of our model using two sets of metrics: Distribution metrics and CAD metrics. For Distribution metrics, we randomly sample 3,000 Sketch-Extrusion (SE) sequences from the generated data and compare them with 1,000 SE sequences from the reference test set. The following values are then computed.

\begin{itemize}
  \item \textbf{Coverage (COV)}: This metric measures the percentage of reference data that has at least one match in the generated data after assigning every generated data point to its closest neighbor in the reference set based on Chamfer Distance (CD).
  \item \textbf{Minimum Matching Distance (MMD)}: This metric represents the average CD between a reference set data point and its nearest neighbor in the generated data.
  \item \textbf{Jensen-Shannon Divergence (JSD)}: JSD quantifies the distribution distance between the reference and the generated data. We convert point clouds into $28^3$ discrete voxels before the practical computation.
\end{itemize}

As for CAD metrics, we simply utilize the same 3,000 SE sequences to compute the following values, as similarly performed in previous work \cite{xu2024brepgen}.

\begin{itemize}
    \item \textbf{F1 Score}: This metric evaluates the predicted extrusions and different primitive types along with their occurrences in the sequences.
  \item \textbf{Novelty}: The percentage of data in the generated set that does not appear in the training set.
  \item \textbf{Uniqueness}: The percentage of data in the generated set that appears only once.
  \item \textbf{Valid Ratios}: The percentage of data in the generated set that are water-tight solids.
\end{itemize}

\section{Additional Qualitative Results}

To further illustrate the generalization ability and geometric coherence of our method,  
we provide extended visual comparisons between GeoFusion-CAD, HNC-CAD, and DeepCAD on the DeepCAD-240 test set, as shown in Fig.~\ref{fig:sup_results}.  
The results cover both short- and long-sequence generation scenarios, including diverse parametric operations such as extrusion, sweep, and multi-sketch modeling.

Specifically, DeepCAD frequently fails to maintain continuity across feature boundaries, resulting in fragmented or disconnected parts, particularly in long-sequence reconstructions.  
HNC-CAD improves on structural connectivity but introduces scaling mismatches and curvature discontinuities in filleted regions.  
In contrast, GeoFusion-CAD accurately reconstructs both high-level topology and fine geometric details, maintaining coherent feature transitions even for highly complex shapes.  
These visual results align with our quantitative findings and highlight the robustness of the proposed geometric state-space diffusion in modeling extended CAD sequences.

\begin{figure*}[t]
    \centering
    \includegraphics[width=0.95 \textwidth]{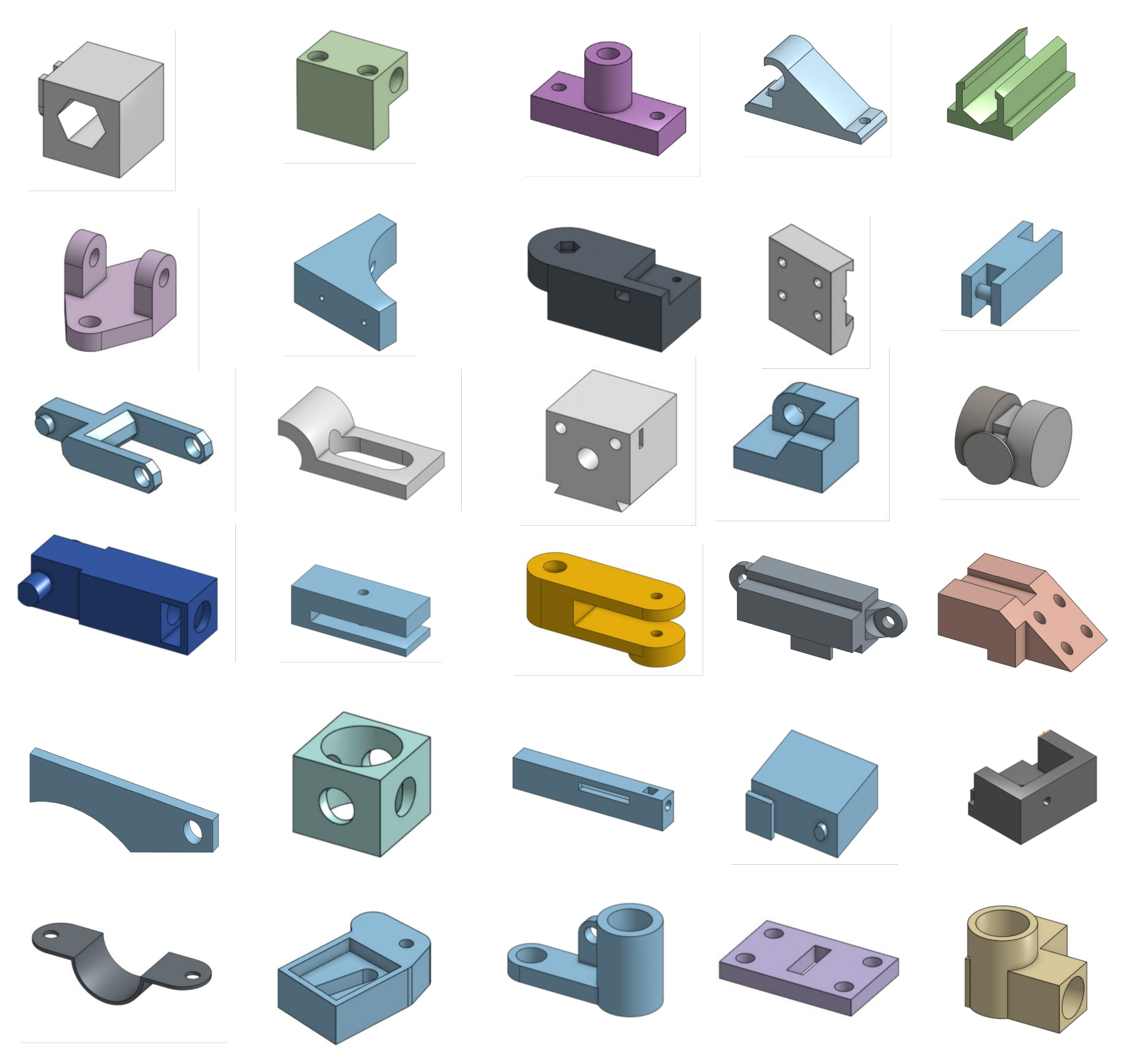}
    \caption{\textbf{Additional CAD modeling results} generated by GeoFusion-CAD.  
    The results cover a variety of industrial-style parts, including prismatic, rotational, and mixed-form geometries.  
    The generated solids are watertight, structurally coherent, and geometrically precise, demonstrating the robustness and generalization of our approach.}
    \label{fig:add_results}
\end{figure*}

\section{Additional Ablation Studies}

This section provides extended analysis on the effects of both the hierarchical tree representation and the G-Mamba diffusion architecture. 
All experiments are performed on the DeepCAD-240 dataset, focusing on long-sequence (40–240 command) generation. 
Metrics include command accuracy (Cmd), parameter accuracy (Param), and geometric distribution measures (COV, MMD, JSD), consistent with the main paper.

\subsection{Effect of Hierarchical Tree Representation}
To investigate the impact of hierarchical encoding, we compare the proposed tree-based representation with a flat sequential formulation in which all sketch–extrusion tokens are arranged in a single, linear sequence (\textit{w/o Tree}). 
As reported in Table~\ref{tab:ablation_tree}, removing the hierarchical structure results in clear degradation across all metrics. 
Command accuracy drops from 91.2 to 87.5 and COV decreases from 73.9 to 69.4, while MMD and JSD both increase notably, indicating weaker geometric alignment. 
This degradation suggests that the hierarchical organization of sketches, faces, and edges is essential for preserving topological dependencies and maintaining global consistency during diffusion.
The hierarchical representation effectively encodes both geometry and topology, allowing the diffusion model to capture long-range dependencies in complex CAD assemblies.



We further analyze the contribution of the proposed G-Mamba block by substituting it with several baseline architectures of comparable capacity:  
(1) a fully-connected multilayer perceptron (MLP),  
(2) a 1D U-Net,  
(3) a Transformer encoder, and  
(4) the original Mamba state-space model.  
Each variant is trained with the same loss functions and optimization schedule as GeoFusion-CAD.

As shown in Table~\ref{tab:ablation_gmamba}, replacing G-Mamba with these alternatives leads to consistent performance degradation, particularly in geometric distribution metrics.  
The MLP version performs the weakest, with high MMD (1.73) and JSD (3.81), indicating unstable geometric diffusion.  
U-Net and Transformer improve slightly but remain inferior to our G-Mamba block, as their attention and convolutional mechanisms struggle to maintain efficiency and long-range coherence over extended command sequences.  
The vanilla Mamba model performs relatively well but still exhibits higher distribution divergence (MMD = 1.19, JSD = 3.05), showing that introducing the geometric state mixer in G-Mamba enhances the model’s ability to propagate geometry-aware features through selective state transitions.  
Overall, the proposed G-Mamba diffusion achieves the best trade-off between computational efficiency and geometric consistency in long-sequence CAD generation.

\section{Additional CAD Modeling Results}

To further demonstrate the generalization and robustness of GeoFusion-CAD,  
we present in Fig.~\ref{fig:add_results} a collection of CAD solids generated unconditionally from random latent diffusion states.  
These examples cover a wide range of geometric and structural configurations, including prismatic, cylindrical, and free-form components with varying topological complexity.  
The generated solids exhibit high geometric fidelity, smooth surface continuity, and consistent feature relationships,  
indicating that the proposed geometric state-space diffusion effectively preserves both local details and global structure during generation.  

Notably, our model successfully produces diverse feature compositions—such as through-holes, fillets, extrusions, and multi-sketch assemblies—without mode collapse or invalid geometry.  
This confirms the scalability of GeoFusion-CAD in capturing complex sketch–extrusion dependencies and supports its potential for future foundation-level CAD generation tasks.


\end{document}